# Optimal Path Planning for Automated Dimensional Inspection of Free-Form Surfaces

Yinhua Liu[a], Wenzheng Zhao[a], Rui Sun[a], Xiaowei Yue[b,1]

[a]*School of Mechanical Engineering, University of Shanghai for Science and Technology, Shanghai, China, 200093*
[b]*Virginia Polytechnic Institute and State University, Blacksburg, VA 24060, USA*

**ABSTRACT**

Structural dimensional inspection is vital for the process monitoring, quality control, and fault diagnosis in the mass production of auto bodies. Comparing with the non-contact measurement, the high-precision five-axis measuring machine with the touch-trigger probe is a preferred choice for data collection. It can assist manufacturers in making accurate inspection quickly. As the increase of free-form surfaces and diverse surface orientations in auto body design, existing inspection approaches cannot capture some new critical features in the curvature of auto bodies in an efficient way. Therefore, we need to develop new path planning methods for automated dimensional inspection of free-form surfaces. This paper proposes an optimal path planning system for automated programming of measuring point inspection by incorporating probe rotations and effective collision detection. Specifically, the methodological contributions include: (i) a dynamic searching volume-based algorithm is developed to detect potential collisions in the local path between measurement points; (ii) a local path generation method is proposed with the integration of the probe trajectory and the stylus rotation. Then, the inspection time matrix is proposed to quantify the measuring time of diverse local paths; (iii) an optimization approach of the global inspection path for all critical points on the auto body is developed to minimize the total inspection time. A case study has been conducted on an auto body to verify the performance of the proposed method. Results show that the collision-free path for the free-form auto body could be generated automatically with off-line programming, and the proposed method produces about 40% fewer dummy points and needs 32% less movement time in the auto body inspection process.

## 1. Introduction

Structural dimensional inspection of auto bodies is necessary for process monitoring, quality control, and variation source diagnosis in the automotive manufacturing industry. In current practice, the high-precision five-axis coordinate measurement machine (CMM) with a touch-trigger probe is one of the most commonly used metrology tools for structural inspection. It has several advantages, such as high measuring accuracy, measurement reliability, low long-term cost, well integration with inspection robotics, programmable path planning, and inspection flexibility for complex product design with free-form surfaces [1-3]. Though the non-contact optical machines are gradually used in the auto body production processes, the high precision CMM with a touch-trigger probe is still considered as the most important pillar in structural dimensional inspection, because it provides high fidelity measurement results which are used as the reference standard for other non-contact measurements and downstream detection, diagnosis and quality control.

For the auto bodies assembled with the free-form sheet metal parts, the main characteristics of dimensional inspection include: (i) a larger number of critical measuring points (MPs), compared to the other products with free-form surfaces (such as blades and biomedical components), are located on the auto body surfaces; (ii) the inspection needs to cope with complex cavity structures and MPs that are located on both outer profiles and inner surfaces of auto body cavities; (iii) the inspection features have diverse normal directions on the inner and outer surfaces, and the inspection should be able to tolerate diverse normal directions of MPs. Therefore, a CMM with a rotary head is required to provide the best inspection directions for various free-form surfaces. However, rotations of the inspection probe introduce new challenges, such as difficult collision avoidance, low measurement efficiency, et al. How to optimize the inspection path planning in order to realize efficient and effective information acquisition is well motivated. In addition, as the new auto body design models are developed rapidly, it poses significant challenges in measuring efficiency and puts forward a considerable demand for automatic path planning strategy of free-form auto bodies.

---

[1] Corresponding author. Tel: +1-540-231-9081, Fax: +1-540-231-3322
Email address: xwy@vt.edu, Postal Address: 1145 Perry Street, Blacksburg, VA 24061-1019, USA.

Many studies have been conducted for optimal path planning with different applications. In the field of path planning of production inspection, we summarize the strategies into three categories: (i) The preliminary path is generated based on heuristic rules, engineering intuition, and guiding hypothesis, i.e., the shortest Hamilton path can be obtained. Then, collision detection, corrections, and verification are carried out in the local paths between MPs sequentially [4-6]; (ii) The collision-free local path between any two MPs is determined at first, and then the global measurement path for all MPs is obtained by the optimization algorithms [7-9]; (iii) the path planning can be generated from tessellated models based on novel Mesh Following Technique (MFT) [10], computer-aided design models and tool models [11], or a spatio-temporal adaptive sampling strategy [12]. The first strategy can generate the executable inspection path directly. However, it may require multiple trials for dummy points' configuration to achieve a collision-free path, which results in low efficiency and inferior solutions for optimization of inspection path planning. For the second strategy, it takes a great amount of time for distance computation of any two measuring points, and focused on minimizing the Hamilton path in global planning with the Traveling Salesman Problem (TSP) solving. Moreover, the aforementioned methods did not incorporate the rotation of the rotary head and stylus, which may contribute to more than 20% of trajectory time for free-form auto body inspection according to current practice. Therefore, these strategies are not sufficient to optimize the path automatically for complex free-form surfaces of novel auto bodies in an efficient manner.

The tolerance-based computer-aided inspection planning system mainly includes collision-free path planning, optimal path generation, and feasible path programming. In terms of collision detection, it can be divided into static collision and dynamic collision detection. Static collision detection has better detection accuracy, while dynamic collision detection can achieve a real-time output of detection results and accepts the structural expansion of measurement machines [13]. As a representative technique in static collision detection, the bounding box method enveloped the probe and the manipulator with a simple geometric structure, and then used the projection crossover algorithm to determine whether and where the collision occurs [14, 15]. The volume bounding method is more suitable to be applied in convex structure objects. However, as the inspection precision increases and the free-form surface becomes more complex, it becomes very challenging for the bounding box to compact the geometric structure. And the computational efficiency in generating the bounding box is also reduced [16]. If the simple structures, such as spheres, axis-aligned bounding box (AABB) and oriented bounding box (OBB), etc. work as the bounding volume, further detections based on these bounding volumes may be needed for achieving collision detection [17]. Besides, methods of space partitioning are also widely used in static collision detection [18, 19]. For example, the octree method can cut the space into many small elements and realize collision detection inside the element body. Overall, it is very challenging for static collision detection algorithms to find a balance between detection accuracy and computational efficiency, especially for the cavity objects with intensive inspection MPs.

As the popularization of intelligent manufacturing in industrial production, dynamic collision detection with real-time outputs becomes increasingly essential. On the one hand, collision detection can be realized by using sensor signals from the perspective of mechanics; On the other hand, virtual simulation of dynamic collision detection can be carried out through spatial geometry calculation [20, 21]. The detection algorithm based on spatial geometry calculation mainly includes three main ideas: (i) the detection process is realized by discretizing the motion trajectory, and the static collision detection is performed at each discrete position. The detection efficiency of these ideas is usually low [22]. (ii) Similar to the bounding box method, the simple geometric structure envelops the space that the measurement element passes by, and the projection crossing is used for detection. This idea becomes very difficult to implement when the trajectory becomes complex [23]. (iii) The motion track of the moving object is expressed by a mathematical formula and then its intersection point with the discretized surface is calculated to realize dynamic collision detection [24-27]. This idea is also called ray tracing technology, which can export accurate results, but the calculation principle is complicated. Existing algorithms are commonly applied in simple geometric structures [28, 29]. However, these algorithms do not work well for accurate dynamic detection of free-form surfaces in real-time, especially for new auto bodies that have intensive MPs, complex cavity structure, and free-form surfaces. When the probe trajectory collides with the measured parts/products, additional space points (also called dummy points) must be added to avoid collisions. Based on the feasible local path, heuristic rules and optimization algorithms used for traveling salesman problem (TSP) solving, including artificial neural network, genetic algorithm (GA), simulated annealing (SA), ant colony optimization (ACO) algorithm, and the dual hierarchical path planning method are used to generate the optimal probe trajectory [30-34]. The aforementioned approaches developed path planning optimization algorithms for programmable tools, and they have been used for different problems in path planning. However, these approaches did not consider the rotations of the probe for free-form surfaces of large size auto bodies, which accounts for a large amount of time during the inspection. In this paper, we propose a systematic dynamic path planning method for automated structural inspection of auto bodies. By incorporating the impact of rotations of the probe on the free-form surface, the proposed method can achieve efficient and high-precision inspection for the free-form surface with intensive MPs in complex cavity structures. Furthermore, a programmed tool is developed to generate the optimal path in an automatic and efficient way. The case study shows that the proposed method needs about 40% fewer dummy points and the CMM movement time is reduced by about 32%.

The paper is organized as follows. Section 2 presents the path planning problem of auto body surfaces. In Section 3, the flowchart of the proposed method is presented firstly. Then, a new collision detection method and a collision-free local path generation method are proposed. In Section 4, considering the probe rotation operations, an inspection time evaluation method in a local path is presented. Then the optimization algorithm can be used to solve the path planning problem based on the inspection time matrix. A case study is conducted on an auto body to illustrate the procedure and performance of the proposed methodology in Section 5. Section 6 summarizes the conclusion.

## 2. Problem statement

The accuracy requirements of auto bodies include shapes, dimensions, and positions with strict tolerances. The types of MPs located on free-form surfaces of new auto bodies become very diverse and complex. For example, inspection routes for multiple MPs, including surface points, holes, slots, trim points, squares and thread-related MPs can be found in Fig. 1. The $i$th MP is denoted by $M_i = (x_i, y_i, z_i, I_i, J_i, K_i)$ $i=1,2,\ldots,m$ in this paper. The set of all the MPs is represented by **M**.

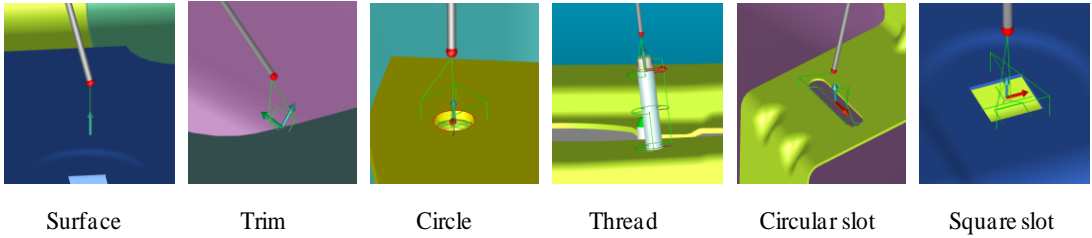

Surface    Trim    Circle    Thread    Circular slot    Square slot

**Fig. 1.** The inspection MPs located on the auto body surfaces.

In order to develop the optimal inspection path planning algorithm, we first analyze the probe movement characteristics. In the feature inspection, the probe movement consists of two parts: the relative high-speed movements between approximate points (APs) of different measuring points and the low-speed movements between APs and MPs. The APs denoted by $Q_i = (x_i, y_i, z_i), i=1,2,\ldots,m$ are located at a distance of $d$ along the normal direction of the MPs. $d$ is also called safety distance. The purpose of setting AP is to avoid collision damage of probe due to auto body manufacturing deviations during low-speed measurements. For example, for a collision-free path, the inspection process from $M_i$ to $M_{i+1}$ can be described in Fig. 2, i.e. the original probe position→$Q_i$→$M_i$→$Q_i$→$Q_{i+1}$→$M_{i+1}$. Thereinto, $Q_i$ and $Q_{i+1}$ are the APs of $M_i$ and $M_{i+1}$, respectively. The dotted blue curve represents the profile of the auto body. $M_i$ is denoted with the red points located in the dotted blue curve, and $Q_i$ is the AP for the corresponding feature. The solid yellow line represents the probe trajectory in the local path between two MPs. The notations in Fig. 2 are also applicable to Fig. 5-9 in this paper.

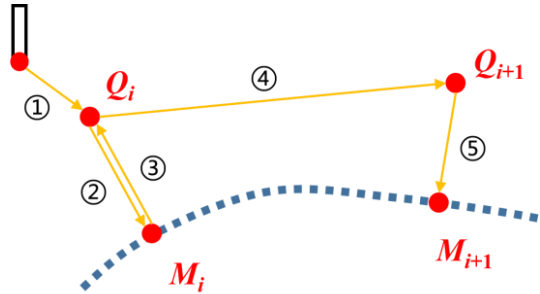

**Fig. 2.** The inspection process of a collision-free path between two MPs.

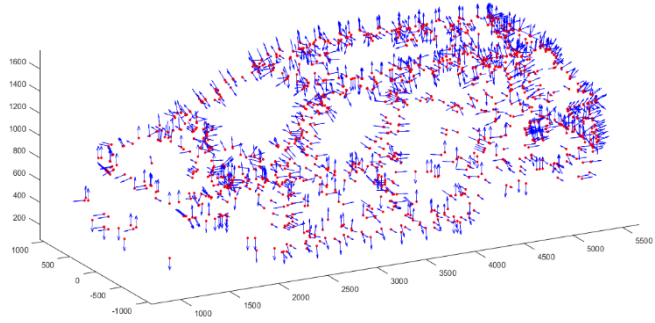

**Fig. 3.** The layout of inspection MPs on an auto-body surface.

Generally, there are hundreds of MPs mentioned above to evaluate the appearance and function of the auto body. Fig. 3 shows the layout of inspection MPs on an auto body. The red dots are the nominal positions of the MPs, and the blue arrows indicate the normal directions of the MPs. In practice, path planning for a large number of MPs is time-consuming, and depends on engineering experience. One path planning may cost weeks for highly skilled engineers, and the generated path is feasible, but non-optimal. Therefore, performing inspection planning automatically and optimally is an important issue in auto body inspection.

The objective of path planning is to output a collision-free path that transverse all the MPs with minimum inspection time efficiently and automatically. Moreover, it needs to consider diverse measurement directions of MPs to ensure high precision. This paper proposes an automatic path planning methodology for free-form auto bodies. Specifically, we develop an effective collision detection method and a collision-free path generation and optimization algorithm by incorporating probe rotations.

The assumptions of this study are listed as follows:
1) Since the measuring time from the AP to MP is unchangeable for any candidate inspection paths, it will be ignored in the total inspection time calculation in this paper.
2) The motion of the probe includes a 3-axis transition and a 2-axis rotation. In the inspection process, only a transition or a rotation of the CMM is executed at one time.
3) During measurement, the angle θ between the stylus direction and the normal direction of the feature is required within a specified range. Otherwise, the stylus should be rotated to a suitable direction to ensure a high-precision measurement.
4) The speed and acceleration rate of the CMM axes are assumed to be constant for the entire inspection process.

### 3. Collision-free local path planning

*3.1. Flowchart of the proposed method*

For the problem of collision-free path planning for auto bodies, the schematic diagram of the proposed methodology is shown in Fig. 4. Based on the auto body CAD database, the inspection MPs on the free-form surfaces are predefined first. Then, the point clouds of the CAD model based on the mesh technique are extracted. The subsequent main procedures include: i) collision detection for the dynamic probe path and component structure, ii) feasible local path generation between feature sets, and iii) inspection time matrix construction and global path planning based on the optimization algorithm. Finally, the optimal path of the CMM is obtained and will be implemented for off-line programming, which ultimately improves the efficiency of auto body inspection.

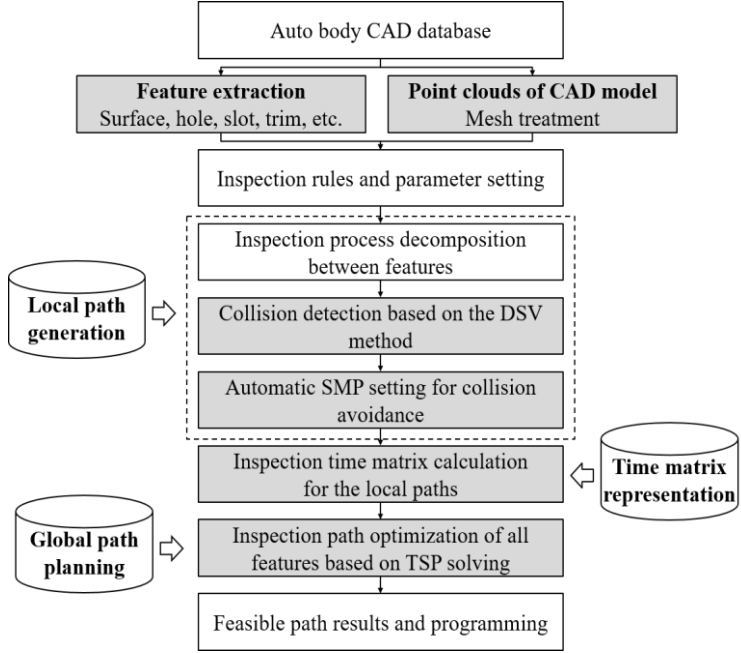

**Fig. 4.** The proposed methodology for collision-free path planning of auto body free-form surfaces.

## 3.2. Probe path collision detection

During the measuring process, the probe of CMM may interfere with the auto body structure. The collisions may lead to inspection inaccessibility of MPs or even equipment damage, so they should be prevented. Therefore, collision detection and automatic avoidance of obstacles have become particularly important. A new dynamic searching volume method will be proposed in this section to realize the automatic avoidance of interference.

The bounding box method is one of the most commonly used methods for collision detection [15]. The principle of the bounding box method is to represent the probes and complex structures with multiple cubes approximately, and then a projection method is used to check the static and dynamic interference in the measurement path. When all the projections in the X, Y, and Z directions do not intersect, it is considered that the probe does not collide with the part structure. Because the part is represented with a large "box", the collision detection may not indicate whether there is real interference between the probe and the part. The sizes of bounding boxes have a large influence on the optimal measurement path, especially for the cases with a large number of MPs. In addition, the bounding box method is applicable for simple solid structures, such as cylinders, vertebras, and cuboids. When it is applied to large-scale free-form structures with external surfaces and internal cavities, the performance of this method is not satisfactory.

In this paper, we propose a new collision detection method called the dynamic searching volume (DSV) method to detect interference in the dynamic CMM inspection path. The method involves calculating the distance from the meshing nodes of the component to the probe path with a predetermined threshold value. The meshing treatment based on a finite element tool is used to auto-mesh the surface of complex structures, as shown in Fig. 5. The element is set as a square and the dimension of the element is denoted by $l$. The free-form surfaces are meshed into discrete nodes for collision detection with the probe path.

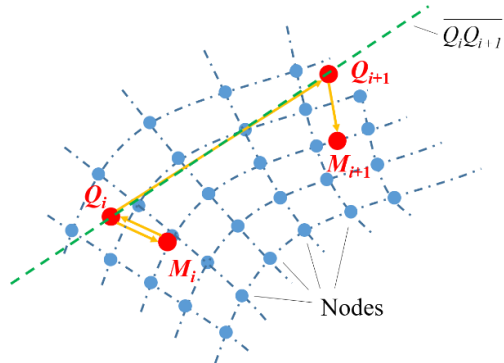

**Fig. 5.** Meshing operation of the auto body structure.

When the CMM probe moves from $M_i$ to $M_{i+1}$, the probe trajectory forms a moving path from the approximate points $Q_i$ to $Q_{i+1}$. According to the coordinates of $Q_i(x_i, y_i, z_i)$ and $Q_{i+1}(x_{i+1}, y_{i+1}, z_{i+1})$, the equation of the line $\overline{Q_iQ_{i+1}}$, which indicates the pass from $Q_i$ to $Q_{i+1}$ in three-dimensional space, can be presented as follows:

$$\frac{x - x_i}{x_{i+1} - x_i} = \frac{y - y_i}{y_{i+1} - y_i} = \frac{z - z_i}{z_{i+1} - z_i} \tag{1}$$

where $(x, y, z)$ denotes the coordinates of the points on the line $\overline{Q_iQ_{i+1}}$.

Because of the large size of the auto body, the number of discrete point clouds is extremely large, usually more than 1 million. To improve the computation efficiency, a dynamic component searching volume method is proposed for collision detection. In the inspection from $Q_i$ to $Q_{i+1}$, the discrete points located in the following constraint area are selected for collision detection of the probe trajectory.

$$\begin{aligned}
\min\{x_{Q_i}, x_{Q_{i+1}}\} - \varepsilon \leq x \leq \max\{x_{Q_i}, x_{Q_{i+1}}\} + \varepsilon \\
\min\{y_{Q_i}, y_{Q_{i+1}}\} - \varepsilon \leq y \leq \max\{y_{Q_i}, y_{Q_{i+1}}\} + \varepsilon \\
\min\{z_{Q_i}, z_{Q_{i+1}}\} - \varepsilon \leq z \leq \max\{z_{Q_i}, z_{Q_{i+1}}\} + \varepsilon \\
i = 1, \ldots, m-1
\end{aligned} \tag{2}$$

where $\varepsilon$ is a non-negative value and is used to determine the searching volume of components for collision detection. $x_{Q_i}$, $y_{Q_i}$, and $z_{Q_i}$ denote the corresponding $x$, $y$, $z$ coordinate of $Q_i$. $m$ represents the number of MPs.

All the nodes in the searching volume are represented by a node set **R**. $\mathbf{R} = \{R_k = (x_k, y_k, z_k) \mid k = 1, 2, \ldots, n\}$. $n$ is the number of nodes in the selected searching volume. The distance $D_k$ between the nodes $R_k$ to the line $\overline{Q_iQ_{i+1}}$ can be calculated. If the condition of $\min\{D_k\} > D_0 (k=1,2,\ldots,n)$ is satisfied, there is no collision for the measurement path from $Q_i$ to $Q_{i+1}$; otherwise, there are interferences in the path. The $D_0$ can be determined based on engineering requirements for inspection safety.

### 3.3. Generation of local path

In the inspection process of auto bodies, the safety distance between $Q_i$ and $M_i$ is determined according to the inspection instrument configurations, usually as 2-5mm. Therefore, collisions usually do not exist in the slow-speed probe moving process between APs and MPs. We mainly focus on collision avoidance between the APs $Q_i$ ($i=1,2,\ldots,m$). When there is no collision in the moving path of the CMM probe, it is not necessary to add spatial movement points (SMPs), also called dummy points. The optimal collision-free path of the probe from the measuring point $M_i$ to the measuring point $M_{i+1}$ can be expressed as

$$Q_i \rightarrow M_i \rightarrow Q_i \rightarrow Q_{i+1} \rightarrow M_{i+1} \rightarrow Q_{i+1}$$

When the probe may have a collision with the component, that means the probe path $\overline{Q_iQ_{i+1}}$ interferes with the component structure, we need to add SMPs so that collisions can be avoided. Taking Fig. 6 as an example, after adding SMPs, a collision-free path from $M_i$ to $M_{i+1}$ can be expressed as

$$Q_i \rightarrow M_i \rightarrow Q_i \rightarrow P_{i(1)} \rightarrow P_{i+1(1)} \rightarrow Q_{i+1} \rightarrow M_{i+1} \rightarrow Q_{i+1}$$

where $P_{i,(1)}$ and $P_{i+1,(1)}$ are the SMPs added in the local path for collision avoidance.

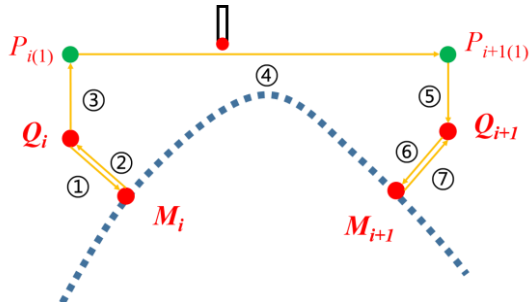

**Fig. 6.** The process of an interference measuring path between MPs.

When a collision occurs in a local measuring path, it is essential to add the SMPs for feasible path generation efficiently. According to the surface orientations of the sequential MPs, the SMPs generation methods can be divided into two scenarios. The first scenario is that the angle $\theta$ between the two normal directions of MPs is less than 180°, and the other is that the angle $\theta$ is equal to 180°, i.e., the directions of two MPs on the sheet metal parts are entirely opposite.

For the first scenario, two rules for SMP generation are proposed to avoid collisions, as shown in Fig. 7(a) and Fig. 8(a), respectively. The main idea of these rules is that the SMPs, i.e. dummy points, are generated along the summation direction of the two normal directions of $M_i$ and $M_{i+1}$. The summation of vectors of $M_i$ and $M_{i+1}$ is represented by

$$V_{i,i+1} = V_i + V_{i+1} = \overrightarrow{M_i Q_i M_{i+1} Q_{i+1}} \tag{3}$$

where $V_i = \overrightarrow{M_i Q_i}$ is the normal direction of $M_i$, i.e., $(I_i, J_i, K_i)$, and $V_{i+1} = \overrightarrow{M_{i+1} Q_{i+1}}$ is the normal direction of $M_{i+1}$, i.e., $(I_{i+1}, J_{i+1}, K_{i+1})$.

Fig. 7 (a) shows a schematic diagram of the process of SMP generation. The initial SMP $P_{i,i+1(0)}$ is the midpoint of $Q_i$ and $Q_{i+1}$. Between the points $Q_i$ and $Q_{i+1}$, the solid line in orange is the path with collision. The new SMP is generated from the position of the last SMP $P_{i,i+1(k)}$ with a distance $h$ along $V_{i,i+1}$, and can be represented in Equation (4)

$$\left| P_{i,i+1(k+1)} - P_{i,i+1(k)} \right|_{V_{i,i+1}} = h \tag{4}$$

The determination of $h$ needs to consider the trade-off between the automatic planning efficiency and the inspection time of the final optimized path, especially for the hundreds of MPs and thousands of meshing nodes of free-form surfaces. In this paper, $h$ is determined based on simulation test and engineering experience (especially the curvature of auto bodies and the density of MPs). The dotted lines in green in Fig. 7(a) is the updated collision-free inspection path. The complete procedure of local collision-free path generation is shown in Fig. 7(b).

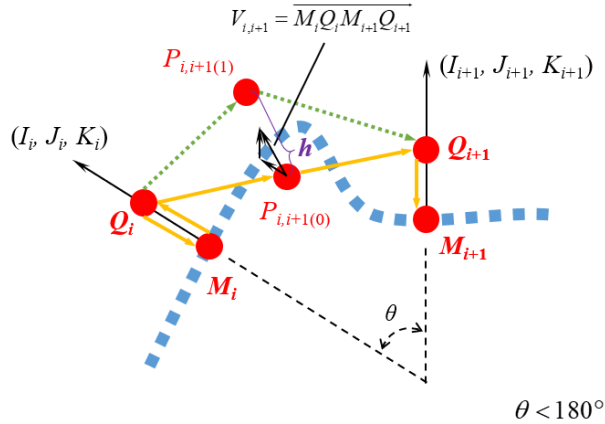

Fig. 7(a). Schematic diagram of the 1st SMP generation rule for scenario 1.

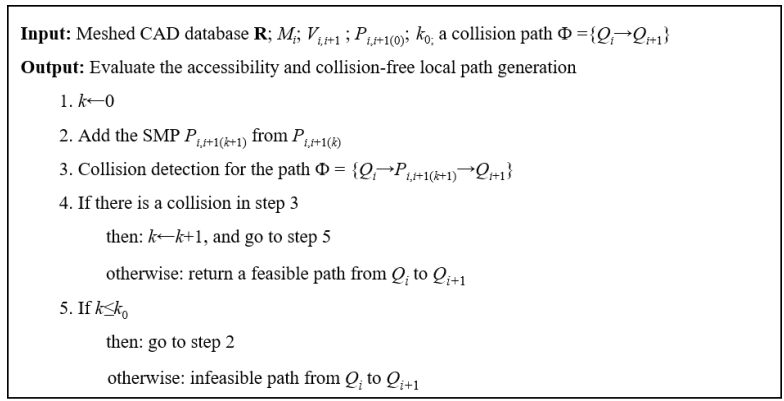

Fig. 7(b). The procedure of the 1st local path generation for Scenario 1.

In Fig. 8(a), collision avoidance is achieved by adding two SMPs with a sequential order from $P_{i(0)}$ and $P_{i+1(0)}$ with a similar way as in Equation (4). Based on several times of SMPs generation, the updated probe path is detected for collision until a feasible path is achieved. For example, for the original collision path $Q_i \to Q_{i+1}$ in Fig. 8(a), the SMPs are generated for 3 times from $Q_i$ and $Q_{i+1}$ sequentially; the corresponding paths are denoted as the solid yellow lines, green dotted ones, and grey dotted ones, which is a collision-free path based on the 2nd rule. The complete procedures of local collision-free path generation are shown in Fig. 8(b).

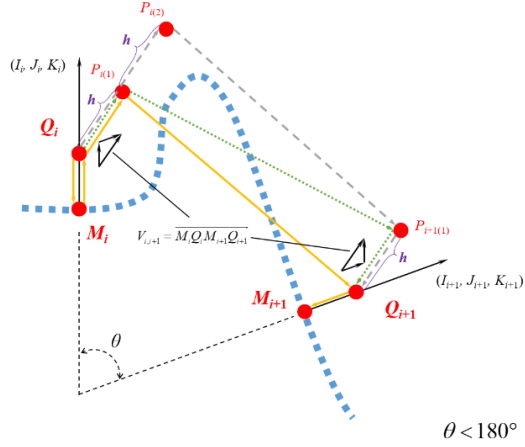

**Fig. 8(a).** Schematic diagram of the 2nd SMP generation rule for scenario 1.

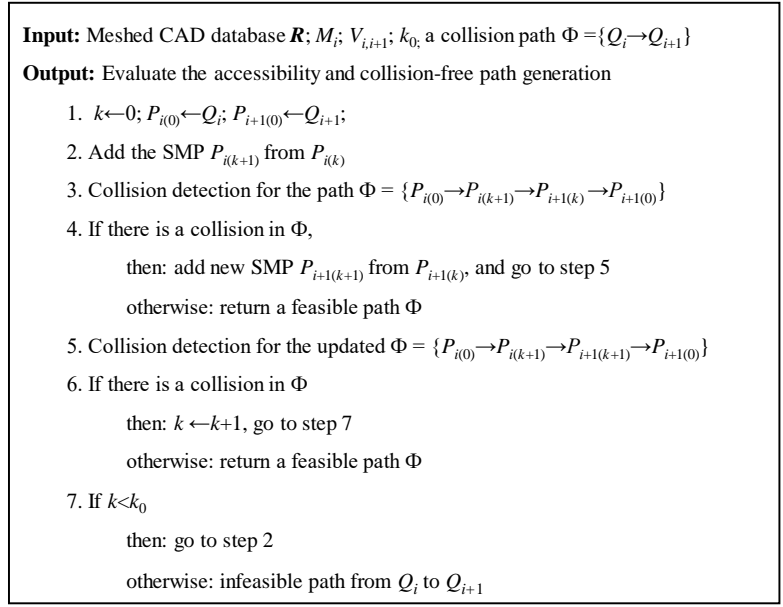

**Fig. 8(b).** The procedure of the 2nd feasible path generation for Scenario 1

Both aforementioned rules are conducted until a local collision-free path is obtained, or the local path is determined as inaccessible. Besides, it is worth mentioning that the generated feasible path from $M_i$ to $M_{i+1}$ may be different from the path from $M_{i+1}$ to $M_i$ due to the complex structures of auto body, rotation sequence of probe, and irregular MPs. At last, the path with less inspection time will be selected as the final collision-free path among MPs.

Some MPs, i.e., $M_i$ and $M_{i+1}$, may have opposite surface orientations, which means the angle between the vector directions is 180°. The positions and normal directions are denoted by $M_i(x_i, y_i, z_i, I_i, J_i, K_i)$ and $M_{i+1}(x_{i+1}, y_{i+1}, z_{i+1}, -I_i, -J_i, -K_i)$. For a path with collision, the SMPs will be generated along with the six vector directions that are perpendicular to the vector $(I_i, J_i, K_i)$. The corresponding 6 direction are $(-J_i, I_i, 0)$, $(-K_i, 0, I_i)$, $(0, -K_i, J_i)$, $(J_i, -I_i, 0)$, $(K_i, 0, -I_i)$, $(0, K_i, -J_i)$, which is simplified as $V_{i,i+1,(u)}$ ($u=1,2,…,6$). Taking the direction $(-J_i, I_i, 0)$ as an example, the SMPs generation process is shown in Fig. 9(a). At first, the SMPs for the two APs are generated along the direction $V_{i,i+1,(u)}$. Then, based on SMPs generation, the updated probe path is detected for collision until a feasible path is achieved, or the maximum number of SMPs is reached. The solid orange lines in Fig. 9(a) is a collision-free inspection path. The complete procedures of local collision-free path generation are shown in Fig. 9(b). The path with less inspection time of different SMP generation direction $V_{i,i+1,(u)}$ will be selected as the final collision-free path among APs.

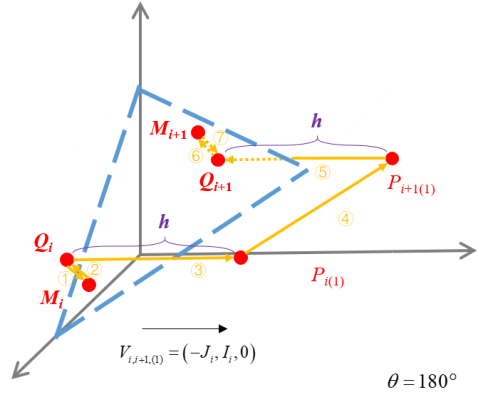

**Fig. 9(a)**. The SMP generation process along $(-J_i, I_i, 0)$

```
Input: Meshed CAD database R; M_i; k_0; a collision path Φ = {Q_i→Q_{i+1}}; V_{i,i+1,(u)} (u=1, 2,..., 6)
Output: Evaluate the accessibility and collision-free path generation
    1. k←0
    2. u←1, P^u_{i(k)} ←Q_i, P^u_{i+1(k)}←Q_{i+1}
    3. Add the SMPs from the two APs along V_{i,i+1,(u)} with a distance h
    4. Collision detection for the path Φ = {Q_i→P_{i(k+1)}→P_{i+1(k+1)}→Q_{i+1}}
    5. If there is a collision in Φ
            then: k←k+1, and add the SMPs with a similar process in step 3
            otherwise: return a feasible path for u
    6. If k ≤k_0,
            then: go to step 4
            otherwise: infeasible path from Q_i to Q_{i+1}
    7. Repeat from step 1-6 for u=2,3,..,6 and return the path with minimum inspection time
```

**Fig. 9(b).** The procedure for feasible path generation for Scenario 2

## 4. Global path planning

*4.1. Inspection time matrix calculation*

Existing studies on path planning can obtain a short path trajectory given a fixed probe direction. They did not consider the potential impacts from the change of probe direction, such as probe rotation angle and probe direction changes after adding SMPs. Probe rotation and direction changes from adding SMPs will have a large impact on the entire inspection time. They may influence the optimization of the inspection path, especially for complex free-form surfaces with a large number of MPs. We propose a new inspection path planning method that incorporates the time from adding SMPs and probe rotations. The measuring process is decomposed into several steps, and the inspection time for all steps is evaluated. Furthermore, the measuring time for any two MPs is calculated, and the inspection time matrix is generated for global path optimization in this section.

For a candidate global inspection path, the inspection time $t$ can be expressed as

$$t = \sum_{i=1}^{m-1} t_{Q_i} + \sum_{i=1}^{m} t_{M_i} \qquad (5)$$

where $t_{Q_i}$ is the total measuring time from the APs $Q_i$ to $Q_{i+1}$, and $t_{M_i}$ is the measuring time from the $i$-th MP to the corresponding AP $Q_i$. For different inspection paths with all required MPs, the summation of all $t_{M_i}$ is a constant given the number of MPs. Therefore, the relevant measuring time for predetermined MPs can be ignored in the optimization of path planning. The objective of path planning is to minimize the probe inspection time between the APs.

Considering the SMPs generation and probe rotation, the local path can be categorized into four scenarios: (a) no SMPs, no probe rotation; (b) with SMPs adding, no probe rotation; (c) with probe rotation, no SMPs; (d) with SMPs adding, with probe rotation.

In the inspection process from $Q_i$ to $Q_{i+1}$, the local inspection time $t_{Q_i}$ includes two parts: the probe transitional time $t^T_{Q_i}$ and the probe rotational time $t^R_i$. According to conditions of collision detection from $Q_i$ to $Q_{i+1}$, $t^T_{Q_i}$ can be calculated as

$$t_{Q_i}^T = \begin{cases} \dfrac{|Q_i - Q_{i+1}|}{v} & \text{if no SMP added} \\[6pt] \dfrac{|Q_i - P_{i(k^i)}| + |P_{i(k^i)} - Q_{i+1}|}{v} & \text{if one SMP added} \\[6pt] \dfrac{|Q_i - P_{i(k^i)}| + |P_{i(k^i)} - P_{i+1(k^{i+1})}| + |P_{i+1(k^{i+1})} - Q_{i+1}|}{v} & \text{if two SMPs added} \\[6pt] A_{inf} & \text{inaccessible path from } Q_i \text{ to } Q_{i+1} \end{cases} \quad (6)$$

where $P_{i(k^i)}$ represents the $k^i$-th SMP in the local collision-free path for $Q_i$, $v$ is the average transitional velocity of the probe, and $A_{inf}$ is a pre-determined large positive value, which means there are no feasible collision-free paths from $Q_i$ to $Q_{i+1}$.

For the local path from $Q_i$ to $Q_{i+1}$, the probe rotational time $t_i^R$ can be calculated as

$$t_i^R = \begin{cases} \dfrac{\theta_i^{A+B}}{\omega} + t_i^s & \text{When probe rotation is needed} \\ 0 & \text{No probe rotation needed} \end{cases} \quad (7)$$

where $\theta_i^{A+B}$ is the angle required to rotate of the probe in the local path from $Q_i$ to $Q_{i+1}$, $\omega$ is the average angular speed of the A, B axis, and $t_i^s$ is the pause time due to probe rotation.

The measuring time from $Q_i$ to $Q_q$ can be denoted by $T_{iq}$ ($i, q=1,2,...,m$). All the values of $T_{iq}$ can be determined and then form an inspection time matrix $\mathbf{T} = \{T_{iq} \mid T_{iq} > 0, i, q=1,2,...,m\}$. The time matrix is a symmetric $m$-order matrix. Based on the time matrix, the global path planning problem can be solved with an optimization algorithm.

### 4.2. Global inspection path optimization

The global path planning determines the sequence of the MPs, including the MPs, APs, and SMPs generated in the local path for collision avoidance, and probe rotation in the inspection process. In this section, the accessibility and measuring time of collision-free path traversing all MPs will be obtained. The CMM measuring time between two APs, including $A_{inf}$ for inaccessible path and specific measuring time of a collision-free local path, can be obtained by the time matrix $\mathbf{T}$. According to the engineering requirement, the probe needs to return to the original position after completing the inspection. Moreover, the inspection path requires each feature to be measured once. The objective of path planning is to obtain a feasible path for all the MPs with the shortest total inspection time.

It is assumed that one of the inspection sequences of the MPs is represented as $\{M_1, M_2, ..., M_m\}$. The corresponding measuring time between every two APs can be obtained based on inspection time matrix $\mathbf{T}$. This problem of finding the collision-free path with the shortest time can be transformed into a TSP. The problem for global path planning can be formulated

$$\min\left\{\sum_{i=1}^{m}\sum_{q=1, q\neq i}^{m} T_{iq} x_{iq} + T_{01} + T_{m0}\right\}$$

$$\sum_{i=1, q\neq i}^{m} x_{iq} = 1, q=1,2,...,m \quad (8)$$

$$\sum_{q=1, i\neq q}^{m} x_{iq} = 1, i=1,2,...,m$$

where $T_{01}$ is the inspection time from the original probe position to the first MP, and $T_{m0}$ is the inspection time from the position of the last MP to the original probe position. $x_{iq}$ is equal to 1 when the path goes from the $i$-th MP to the $q$th MP. Otherwise, $x_{iq}$ is set to be 0.

The TSP is a combinatorial optimization problem, and it is NP-hard. The prior studies on this problem include branch and bound, linear programming and dynamic programming methods. However, with an increase in the number of targets, the computation of a feasible path increases exponentially, and it is difficult to obtain the global optimal solution. Different heuristic algorithms have been developed for TSP, including SA, GA, ACO, A-Star algorithm etc. [30-34]. With the advantage in general flexibility in TSP solving, the SA, GA, ACO algorithms were chosen for comparative analysis, and the final performance of the optimal inspection path will be shown in the case study.

## 5. Case study

In order to verify the effectiveness of the proposed method, we illustrate it with model simulation and a case study on auto body inspection. Automatic metrology was used to measure the deviations of an auto body in the assembly line. Two horizontal-cantilever CMMs were used to collect the data of dimensional deviations of the auto body. In order to improve the measurement efficiency, the plane XOZ in the auto body coordinate system is used to divide the measurement characteristics into the left and right parts, which will be measured by the mentioned two horizontal-cantilever CMMs respectively. In this case, we focus on path planning and optimization on the left side of the auto body. The inspection path planning on the other side of the body can be performed similarly.

According to the features predefined in the design stage, there are total of 543 MPs on the left side of the auto body, including 360 surface points, 34 trim points, 88 holes, 14 threaded columns, 30 circular slots, and 17 square slots. Each MP corresponds to different spatial coordinates and normal directions. For the CAD model of the auto body, the finite element tool was used to mesh the free-form surfaces. The parameters used in this case are shown in Table 1.

**Table 1** The parameters used in the case study

| Parameter | $l$ (mm) | $D_0$ (mm) | $d$ (mm) | $h$ (mm) |
|---|---|---|---|---|
| value | 4 | 4 | 5 | 10 |
| Parameter | $k_0$ | $\omega$ (°/s) | $t_i^s$ (s) | $v$ (mm/s) |
| value | 10 | 1 | 0.3 | 85 |

*5.1. Model simulation*

According to the procedure in Fig. 4, the meshed nodes of the CAD model, the coordinates, and normal directions of APs for all MPs are taken as the inputs for path planning. The path between two APs is checked with the proposed DMV collision detection algorithm. When a potential collision may occur, SMPs are added automatically to obtain a collision-free local path between two MPs. In the situation that there is no feasible local path, or the measuring time of a feasible local path is greater than 200 seconds, the measuring time is denoted by $A_{inf}$. Such configuration can avoid being stuck in a non-feasible local path and significantly increasing the inspection time. Finally, the inspection time matrix T is calculated, i.e., the time required for the collision-free path between two APs can be obtained. Partial results of the matrix by the 543 MPs are shown in Table 2. The value of $T_{iq}$ represents the total time of the local collision-free path from $Q_i$ to $Q_q$. T is usually a symmetric matrix, which can be corrected by the method of "path inversion" when the asymmetry is caused by the obstacle avoidance strategy.

**Table 2** Part of the inspection time matrix for the 543 MPs

| No. | 1 | 2 | 3 | 4 | 5 | 6 | 7 | 8 | …… | 543 |
|---|---|---|---|---|---|---|---|---|---|---|
| 1 | 0 | 29.77 | 30.00 | 30.70 | $A_{inf}$ | 33.49 | 31.35 | 31.29 | | 32.82 |
| 2 | 29.77 | 0 | $A_{inf}$ | 1.56 | 1.99 | 3.61 | 3.12 | 4.20 | | 3.97 |
| 3 | 30.00 | $A_{inf}$ | 0 | 0.50 | 0.91 | 3.01 | $A_{inf}$ | 4.28 | | $A_{inf}$ |
| 4 | 30.70 | 1.56 | 0.50 | 0 | 0.45 | 2.69 | 4.53 | 4.70 | | $A_{inf}$ |
| 5 | $A_{inf}$ | 1.99 | 0.91 | 0.45 | 0 | $A_{inf}$ | 4.57 | 5.08 | | $A_{inf}$ |
| 6 | 33.49 | 3.61 | 3.01 | 2.69 | $A_{inf}$ | 0 | 4.12 | 5.01 | | $A_{inf}$ |
| 7 | 31.35 | 3.12 | $A_{inf}$ | 4.53 | 4.57 | 4.12 | 0 | 0.90 | | 3.24 |
| 8 | 31.29 | 4.20 | 4.28 | 4.70 | 5.08 | 5.01 | 0.90 | 0 | …… | 4.17 |
| …… | | | | | | | | | …… | …… |
| 543 | 32.82 | 3.97 | $A_{inf}$ | $A_{inf}$ | $A_{inf}$ | $A_{inf}$ | 3.24 | 4.17 | …… | 0 |

Based on the feasible local path time between any two APs, the total measurement time of the probe traversing all the critical MPs can be calculated. In this study, the algorithms based on SA, ACO and GA are used to optimize the probe path, and their results were compared.

With the parameter tuning for the optimization algorithms, Table 3 shows the optimal results for path planning with different algorithms. The calculation was implemented on a PC with a Core(TM) i3-2350M 2.30 GHz CPU and 4.00 GB RAM. The results show that the SA algorithm is superior to the other two algorithms in terms of both accuracy and calculation efficiency. In terms of calculation efficiency, the SA and GA methods tend to obtain better performance with much less computation time. In terms of path planning results, the measuring time of the optimized path based on SA was 68.9% of that based on ACO, and 44.6% of that based on GA. Therefore, the optimal path of the SA algorithm was selected as the final inspection path of the 543 MPs.

**Table 3.** Results comparison based on different optimization algorithms

| Algorithm | SA | ACO | GA |
|---|---|---|---|
| Calculation time ($s$) | 220.8 | 1219.3 | 349.3 |
| Measuring time of optimized path($s$) | 1282.22 | 1860.56 | 2874.90 |

Based on the SA algorithm, inspection path planning and optimization algorithm were carried out for 538 of the 543 MPs, and the remaining 5 inaccessible MPs are identified. For these 5 MPs, no feasible inspection path was found to meet the measurement requirements in Table 1, and they need further investigation (e.g., optimizing inspection feature design according to engineering domain knowledge). A total of 145 SMPs, 65 probes rotations and 739 segments of probe translation motion are included in the final optimized detection path. The optimized inspection path is visualized in Fig.10. The blue scatter is the auto body structure. The 543 measurement MPs on the auto body are highlighted with red dots, and the optimized probe trajectory is displayed in the yellow line. The green star means the stylus rotation during the measuring process. The generated SMP in the local path is denoted with a black circle. The proposed method can effectively conduct collision detection, local path generation, and also global path optimization for the critical MPs, which can enable automatic quality inspection and further process optimization.

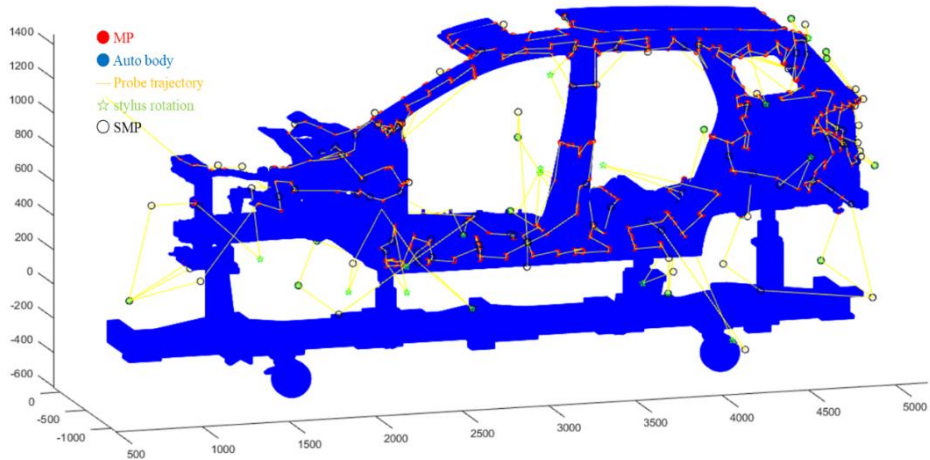

**Fig. 10.** Schematic diagram of the optimal inspection path based on SA algorithm.

*5.2. Comparative analysis and verification*

To validate the proposed path planning method, several experienced engineers from an industrial partner program the inspection path based on commercial software, PC-DMIS. Then the accessibility of the feature and the effectiveness of inspection path planning are evaluated. The final optimal path is applied to real auto body inspection. The inspection path results of the model simulation are also imported into PC-DMIS software for comparative analysis, and the accessibility and effectiveness of the proposed method are verified. The result of the simulation is shown in Fig. 11. The lines denote the probe trajectory during the inspection process. Correspondingly, the green lines represent a collision-free path, and the red lines represented the path with collision.

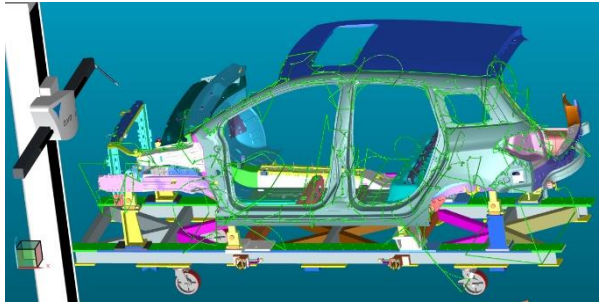

**Fig. 11.** Verification of the optimal inspection path for collision detection

The inaccessible points generated from the model simulation are consistent with those obtained from multiple experiments in practice. Furthermore, the proposed optimal global path is compared with the current path in the engineering practice. The comparison results are shown in Table 4. The improvement rate was used to evaluate the proposed method, and it is the ratio between the saved inspection time and the inspection time needed in traditional planning. As shown in Table 4, the inspection time spent on probe transitions was reduced by 29% by applying the new inspection path planning.

**Table 4** Comparative analysis in engineering practice

| item | Proposed automatic planning | Traditional path Planning | Improvement rate |
| --- | --- | --- | --- |
| Probe transition (s) | 1022 | 1448 | -29% |
| Probe rotation (s) | 260 | 240 | 8% |
| Probe replacement (s) | 104 | 364 | -71% |
| Total inspection time (s) | 1386 | 2052 | -32% |
| Total number of probe direction | 25 | 35 | -29% |
| Number of probe rotations | 65 | 60 | 8% |
| Number of SMPs | 145 | 284 | -49% |
| Inaccessible MPs | 5 | 5 | 0 |

Based on the comparison, the proposed inspection path planning can give superior results. Specifically, the probe transition is reduced by 29%, and the number of SMPs is reduced by 49%, which would enhance the stability of the measurement system and improve inspection efficiency. Overall, the total inspection time for all the MPs has been reduced by 32% for each auto body.

## 6. Conclusion

This paper presented a systematic framework for an inspection path planning method for free-form auto body surfaces. The proposed methodology includes: (*i*) a dynamic searching volume method for collision detection and a rule-based collision-free local path generation method; (*ii*) an inspection time matrix representation considering probe translation, rotation and pause time; and (*iii*) a SA-based optimization algorithm for inspection path planning. The case study shows the effectiveness and efficiency of the proposed method. Results show the collision detection method is effective compared to the current commercial software. Based on the proposed method, the probe movement time between the APs is reduced by 32%, and the number of SMPs is reduced by 49%. Moreover, this methodology provides an automatic scheme for optimal path planning of auto body inspection. In summary, the proposed path planning method can be used for off-line programming by automobile manufactures. The optimal path planning and automatic inspection can be enabled to improve inspection efficiency significantly. The proposed method can not only be applied to auto body inspection planning but also be extendable to other complex structures like aircraft and train bodies. Since the optimal path planning is to obtain the maximum information with the least cycle time, it has similar characteristics with active learning [35]. For the future work, mathematical connection between optimal path planning and active learning will be explored. Moreover, potential path planning methodology will be investigated from the information theory and active learning perspective.

**Acknowledgments**

This project was partly supported by the National Natural Science Foundation of China (Grant No. 51875362) and partly supported by Shanghai General Motors Corporation.